\documentclass[10pt,twocolumn,letterpaper]{article}

\usepackage{iccv}
\usepackage{times}
\usepackage{epsfig}
\usepackage{graphicx}
\usepackage{amsmath}
\usepackage{amssymb}
\usepackage{multirow}
\usepackage{subfigure}
\usepackage[vlined,boxed,commentsnumbered, ruled, linesnumbered]{algorithm2e}
\usepackage{mathrsfs}
\usepackage[breaklinks=true,bookmarks=false]{hyperref}

\iccvfinalcopy 

\def\iccvPaperID{9580}

\ificcvfinal\fi

\begin{document}

%%%%%%%%% TITLE
\title{Adversarial Training with Natural Transformation}

\author{Shuo Wang, Surya Nepal, Marthie Grobler, Kristen Moore\\
CSIRO\\
Melbourne, Australia\\
{\tt\small shuo.wang@csiro.au}
% For a paper whose authors are all at the same institution,
% omit the following lines up until the closing ``}''.
% Additional authors and addresses can be added with ``\and'',
% just like the second author.
% To save space, use either the email address or home page, not both
\and
Lingjuan Lyu\\
National University of Singapore\\
Singapore\\
{\tt\small lyulj@comp.nus.edu.sg}
}

\maketitle
% Remove page # from the first page of camera-ready.
\ificcvfinal\thispagestyle{empty}\fi

%%%%%%%%% ABSTRACT
\begin{abstract}
Previous robustness approaches for deep learning models such as data augmentation techniques via data transformation or adversarial training cannot capture real-world variations that preserve the semantics of the input, such as a change in lighting conditions. To bridge this gap, we present NaTra, an adversarial training scheme that is designed to improve the robustness of image classification algorithms. We target attributes of the input images that are independent of the class identification, and manipulate those attributes to mimic real-world natural transformations (NaTra) of the inputs, which are then used to augment the training dataset of the image classifier.
Specifically, we apply \textit{Batch Inverse Encoding and Shifting} to map a batch of given images to corresponding disentangled latent codes of well-trained generative models. \textit{Latent Codes Expansion} is used to boost image reconstruction quality through the incorporation of extended feature maps. \textit{Unsupervised Attribute Directing and Manipulation} enables %to find latent directions towards specific attribute change, and then achieve interpretable manipulations for generating natural transformations. 
identification of the latent directions that correspond to specific attribute changes, and then produce interpretable manipulations of those attributes, thereby generating natural transformations to the input data.
We demonstrate the efficacy of our scheme by utilizing the disentangled latent representations derived from well-trained GANs to mimic transformations of an image that are similar to real-world natural variations (such as lighting conditions or hairstyle), and train models to be invariant to these natural transformations. 
Extensive experiments show that our method improves generalization of classification models and increases its robustness to various real-world distortions.

\end{abstract}

%%%%%%%%% BODY TEXT
\section{Introduction} 
%The principle by which neural networks are trained to minimize their average error on the training data is known as Empirical Risk Minimization (ERM) [1]. ERM has, for the most part, enabled breakthroughs in a wide variety of fields [2–4], and this success has lead to the usage of neural networks in applications that are safety-critical [5]. ERM, however, is only guaranteed to produce meaningful models Figure 1.  A model obtained through classical training classifies the same face as both “smiling” and “not smiling” (depending on the variations). Our model remains consistent in terms of classification. Note that these persons “do not exist” and have been generated using a StyleGAN model. when the data encountered during training and deployment is drawn independently from the same distribution. 

\begin{figure}[!htb]
\centering
\subfigure[The original image is at the top left; the rest of the images are misclassified variations under various angles of light irradiation.]{
\begin{minipage}{0.5\textwidth}
\includegraphics[width=\textwidth]{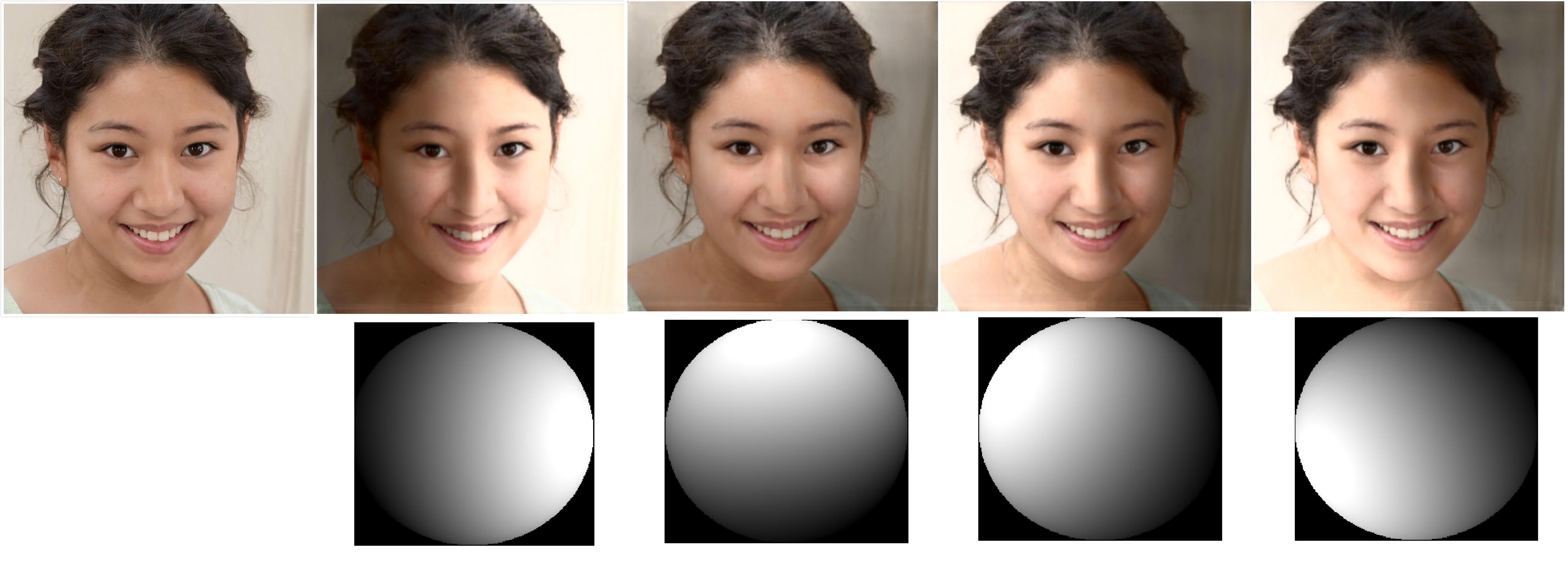} 
\end{minipage}
}
%\vspace{-0.2cm}
\subfigure[The original image is at the top left; the rest of the images are misclassified variations, such as hairstyle, face orientation, exposure condition, background, visual orientation, eyebrows, etc.]{
\begin{minipage}{0.5\textwidth}
\includegraphics[width=\textwidth]{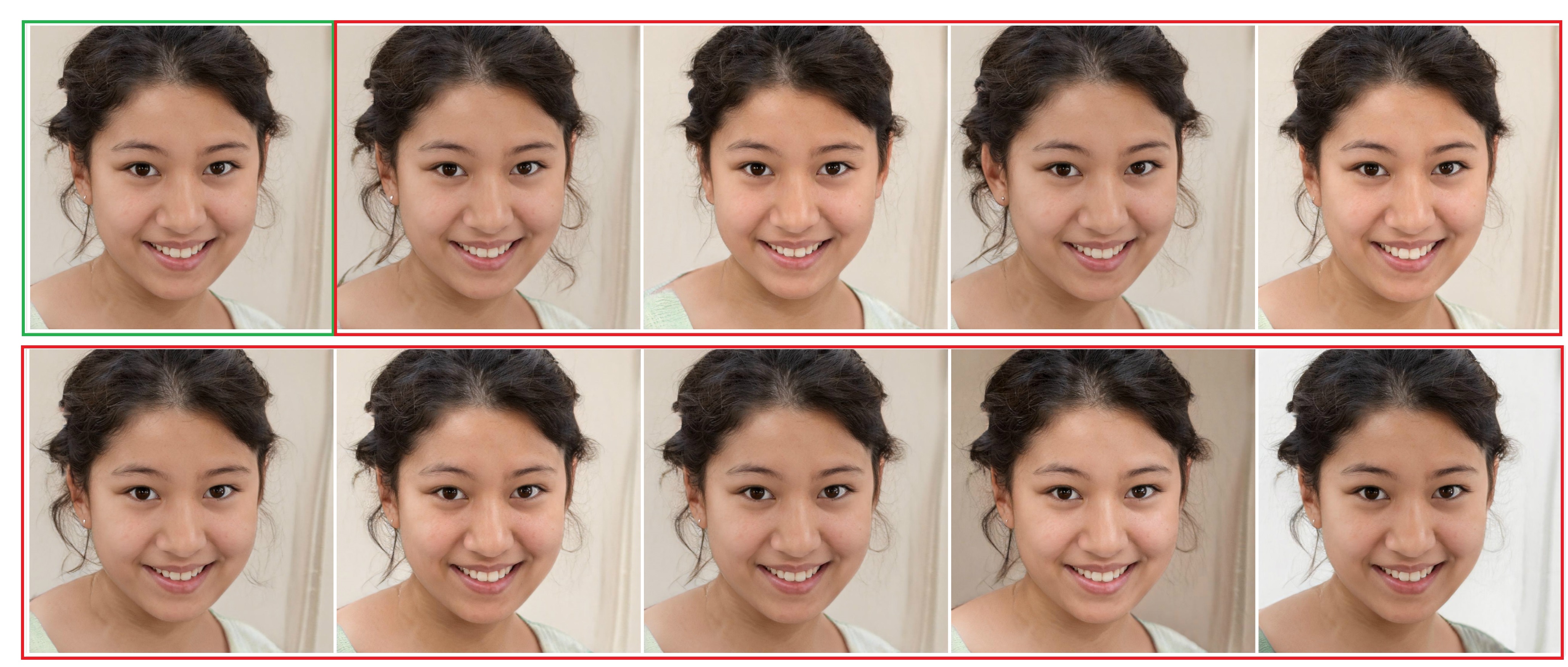} 
\end{minipage}
}
%\vspace{1mm}
\caption{VGGFace model can be deceived by image transformations that mimic real world distortions.}
\label{fig:1}
\end{figure}

The success of adversarial attacks has demonstrated the susceptibility of deep learning models to making incorrect predictions with high confidence, due to small but carefully chosen deviations to the input, i.e. adversarial perturbations  \cite{carlini2017towards, goodfellow2014explaining, kurakin2016adversarial, szegedy2013intriguing}.
Figure~\ref{fig:1} shows that even small variations in the angle of light irradiation and face rotation, background, and exposure conditions for images can result in drastic degradation in the performance of well-trained deep neural networks. 

These examples reveal that state-of-the-art deep learning models sometimes fail to generalize to small variations of the input. 
A possible reason for this is that the representations learnt by them are not good and the robustness of the learned deep neural networks is limited. If deviations exist between training and testing instances, it is commonplace for models to fail in catastrophic ways. 
% \begin{figure}[!htb]
% 	\centering
% 	\setlength{\abovecaptionskip}{-0.05cm}
% 	\setlength{\belowcaptionskip}{-0.2cm}
% 	\includegraphics[width=3.5in,height=2in]{figure1.jpg}
% 	\caption{The demonstration of limited robustness of deep facial recognition model, due to rotation, lighting condition, color change, etc.}
% \end{figure}

%\vspace{-2mm}
%%\vspace{-3mm}

On the one hand, approaches to improve robustness have been found to fall short of their goals, in particular data augmentation techniques via data transformation. Transformations such as cropping, flipping, scaling, color jittering, and region masking (Cutout) are commonly used augmentations for vision models. Training on the corrupted data only forces the memorization of such corruptions, and as a result these models fail to generalize to new corruptions \cite{vasiljevic2016examining,geirhos2018generalisation}. Works such as Mixup \cite{zhang2017mixup} or AutoAugment \cite{cubuk2018autoaugment} pave the way to further improvements, but still require intricate fine-tuning to succeed in practice. Furthermore, these transformations tend to destroy the image semantics.

On the other hand, prior works \cite{goodfellow2014explaining, papernot2016distillation, kannan2018adversarial, xie2019feature} have shown that adversarial training \cite{madry2017towards} is effective in building models that are robust to adversarial perturbations. 
Given a perturbation threshold, $\epsilon$, adversarial
training intends to minimize the loss of robustness, the worst-case loss within the $\epsilon$-ball around each example, leading to a min-max optimization problem. However, such perturbation fails to capture real-world variations that preserve the semantics of the input, such as a change in lighting conditions. 
%The existence of imperceptible perturbations that alter a model’s output, demonstrates that supervised learning algorithms still fail to capture the true causal relationships between signal and label. The degradation of performance that occurs when shifting between training and adversarial (or otherwise corrupted) distributions indicates that neural networks pick up on correlations that are not necessarily robust to small input perturbations [23]. The existence of imperceptible adversarial perturbations highlights just one form of spurious correlation that causes undesirable behaviors in the networks we train. 
%However, applying data augmentation to GANs is fundamentally different. If the transformation is only added to the real images, the generator would be encouraged to match the distribution of the augmented images. As a consequence, the outputs suffer from distribution shift and the introduced artifacts. Alternatively, we can augment both the real and generated images when training the discriminator; however, this would break the subtle balance between the generator and discriminator, leading to poor convergence as they are optimizing completely different objectives. 

This paper focuses on training models to be more robust to natural transformations that mimic real world distortions while preserving the semantic attributes of input images, such as variations to the subject's face orientation, degree of eye opening, or the perspective orientation of objects in Figure~\ref{fig:2}. % that preserves the semantic attributes of the image. 
Instead of the conventional data augmentation and adversarial training on $L_p$-norm bounded perturbations, we achieve this robustness via controllable attributes editing on given high-fidelity inputs. 
%In particular, we address the question: “Given a generative model with a sufficiently good disentangled representation that aligns well with the perturbations of interest, can we train neural networks that are resistant to bias and spurious correlations present in the training data?” 

%\vspace{-3mm}
\begin{figure}[!htb]
	\centering
	\includegraphics[width=3.5in,height=3.0in]{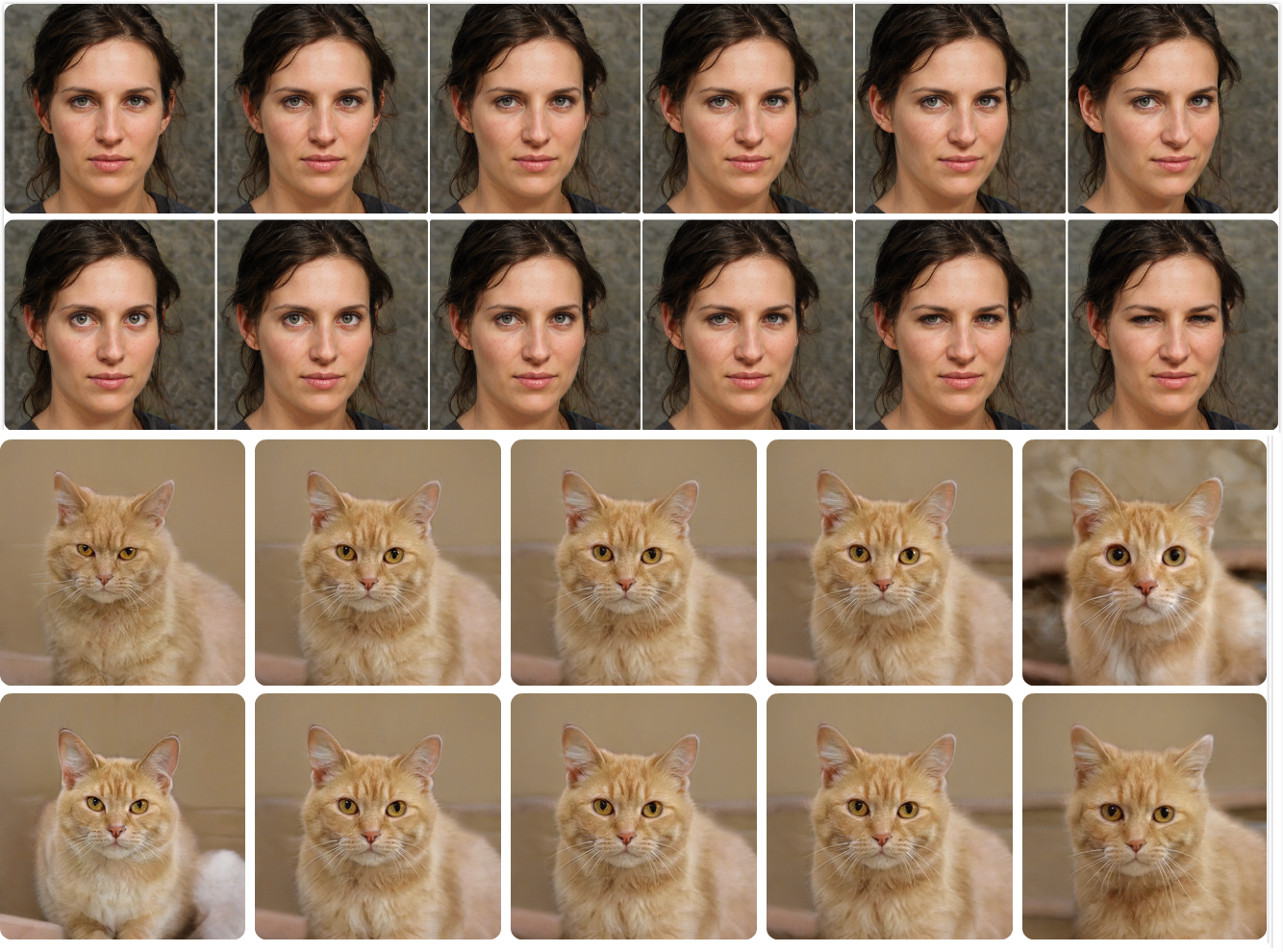}
	\caption{The demonstration of natural transformations that mimic four real-world  variations, i.e. face orientation and eye-opening degree for human, angle of elevation and body orientation for cat.}
\end{figure}
\label{fig:2}
%\vspace{-2mm}

In terms of generating high-fidelity images, Generative Adversarial Networks (GANs) have paved a feasible path towards generating high-fidelity images and producing a disentangled latent representation with a rich linear structure, e.g. StyleGAN \cite{karras2019style, karras2020analyzing} and BigGAN \cite{ brock2018large} . 
However, since the standard GAN model was initially designed for synthesizing images from random noises, applying trained GAN models to real image post-processing remains challenging. 
Existing methods commonly invert a given image back to the latent space either by back-propagating one by one, or by learning an additional encoder. However, the efficiency and reconstruction quality from both of these methods are far from ideal.
The main challenges in the manipulation of selected attributes of given images while preserving the other details include:
\textit{(1) obtaining an efficient reverse function to map massive images to the disentangled latent space of a well-trained generator model, especially for high resolution image generators;} 
\textit{(2) recovering as many details as possible of any arbitrary real image in the reversed latent codes using all the possible composition knowledge learned in the deep generative representation;} 
\textit{(3) finding a proper latent direction that corresponds to the latent codes, that changes only the desired attribute, and to do so with minimum supervision, or ideally in an unsupervised manner.} 
This work address these challenges, and we summarise our contributions as follows: 
% \llv{ the previous 3 points sound more likely strategies, rather than contributions?
\begin{enumerate}
\item We use \textit{Batch Inverse  Encoding and Shifting} (BIES) to map massive given images to the corresponding disentangled latent codes of well-trained generative models at one time, which enables the inverse function of the generator to be accurately learned via an encoder in an end-to-end and unsupervised manner. 
\item We adopt \textit{Latent Codes Expansion} (LCE) to boost image reconstruction quality, and apply \textit{Unsupervised Attribute Directing and  Manipulating} (ADM) to find latent directions towards specific attribute change and then obtain interpretable manipulation for generating nature transformations. 
\item We develop a new framework to leverage the disentangled latent codes of a well-trained generative model to generate a set of  \textit{natural transformation} ((NaTra)) copies for each given image. This enables us to improve robustness via adversarial training on the natural data augmentation in a purely data-driven fashion.
\item We conduct extensive experiments on CelebA-HQ, LSUN datasets, and compare NaTra augmentation with the conventional data augmentation to demonstrate under which conditions NaTra achieves higher accuracy. We empirically demonstrate that accuracy is not necessarily at odds with robustness, once we consider natural variations other than $L_p$-norm bounded variations. 
\end{enumerate}    

\section{Related work}
\subsection{StyleGAN}
StyleGAN is a generator architecture for generative adversarial networks proposed by Karras et al. \cite{karras2019style} and improved in \cite{karras2020analyzing}.  %In this work, we rely on the style mixing property. 
Formally, the StyleGAN architecture is composed of two stages. The first stage takes a latent variable $z \sim N(0, 1)$ that is not necessarily disentangled and projects it into a disentangled latent space $w = map(z)$. The second stage synthesizes an image $x$ from the disentangled latents $w$ using a decoder $x = dec(w)$. 
%Overall, the process of generating an image x using a StyleGAN network is defined as 
% \begin{equation}
%     x=dec \circ map(w)~where~z \sim N(0,1)
% \end{equation}
The intermediate latent variable $z$ provides some level of disentanglement that affects image generation at different spatial resolutions which allows us to control the synthesis of an image. 
%In particular, we can apply the “style” of an image to another by mixing the disentangled latents of these images together. In the context of face generation, the styles corresponding to coarse spatial resolutions affect high-level aspects such as pose, and styles of fine resolutions mainly affect the color scheme. In the rest of this manuscript, we focus on variations of the finer style. %Concretely, our experiments in Section 5 assume that the fine attributes $z_n$ are label independent, while the coarse attributes $z_c$ may be label dependent. Consequently, the finer style $z_{B,n}$ of an image $x_B$ can be applied to another image $x_A = dec(z_{A,c}, z_{A,n})$ via $dec(z_{A,c} , z_{B,n})$. 
%Figure 5b shows a nominal image and two variations of that image obtained by mixing the finer style of two other images. 

To reverse the generation process, existing approaches fall into two types. One is to directly optimize the latent code by minimizing the reconstruction error through back-propagation \cite{creswell2018inverting, lipton2017precise, abdal2019image2stylegan}. The other is to train an extra encoder to learn the mapping from the image space to the latent space \cite{bau2019inverting, zhu2016generative, perarnau2016invertible, luo2017learning}. However, the reconstructions achieved by both methods are far from ideal, especially with high resolution images.
%%%%%%%%%%%%%%%%%%%%%%%%
\iffalse
\subsection{Data augmentation} 
Data augmentation can reduce generalization errors. For image classification tasks, random flips, rotations and crops are commonly used for data augmentation \cite{he2016deep}. More sophisticated techniques such as Cutout \cite{devries2017improved} (which produces random occlusions), CutMix \cite{yun2019cutmix} (which replaces parts of an image with another) and mixup \cite{zhang2017mixup} (which linearly interpolates between two images) all demonstrate extremely compelling and surprising results. While these methods often result in images that are visibly corrupted and void of semantic meaning (even to the human eye), the resulting models often achieve state-of-the-art accuracy across a wide range of datasets. %Figure 2 (a) and (b) demonstrate a comparison of these different techniques.
Some of these data augmentation techniques have been applied to latent representations of the input (rather than the input itself) \cite{verma2019manifold}. However, these techniques did not %focus on 
consider the effect of data bias. 
\fi
%%%%%%%%%%%%%%%%%%%
\subsection{$L_p$-norm perturbations and Adversarial robustness beyond $L_p$-norm%and adversarial training.
} 
Generating pixel-level adversarial perturbations has been and remains extensively studied \cite{moosavi2019robustness,goodfellow2014explaining,szegedy2013intriguing, madry2017towards, papernot2016distillation}. Most works focus the robustness of classifiers under $L_p$-norm bounded perturbations. In particular, it is expected that a robust classifier should be invariant to small perturbations in the pixel space (as defined by the $L_p$-norm). %Recent works \cite{qin2019adversarial, zhang2019theoretically} continue to find novel approaches to enhance robustness. 
However, robustness to semantically meaningful perturbations remains a largely unexplored problem. 

% \subsection{.} 
\cite{engstrom2017rotation} and \cite{kanbak2018geometric} explored geometric transformations such as rotations and translation of images. Early works (e.g., Baluja and Fischer \cite{baluja2017adversarial}) also demonstrated that it is possible to go beyond analytically defined variations by using generative models to create perturbations. \cite{song2018constructing,xiao2018generating} used a pre-trained AC-GAN \cite{odena2017conditional} to generate perturbations; and they demonstrated that it is possible to generate semantically relevant perturbations for tasks such as MNIST, SVHN and CELEBA. Lastly, \cite{qiu2019semanticadv} have attempted to generate adversarial examples by interpolating through the attribute space defined by a generative model. With the exception of \cite{jalal2017robust}, in which the authors strongly limit semantic variations by keeping the perturbed image close to its original counterpart, there has been little to no work demonstrating robustness to large semantically plausible variations. As such, the effect of training models to be robust to such variations is unclear. To the best of our knowledge, this paper is the first to analyze the difference between adversarial training and data augmentation in the space of semantically meaningful variations. 

\section{Robustness via Adversarial Training on Natural Transformation}
\subsection{Problem Definition}
Adversarial training has become one of the most effective methods for improving robustness of neural networks.
Adversarial training in this study can be considered as a data augmentation technique that fine-tunes DNNs on additional natural transformations of natural inputs to improve the robustness of the classifier to feature variations.
Essentially, we implement an autoencoder scheme to conduct such real-world natural transformations. \textit{Batch  Inverse  Encoding and Shifting -- BIES} is used to map massive given images to corresponding disentangled latent codes of well-trained generative models at one time, i.e. $enc: X \rightarrow Z$. The well-trained generator can be used as a decoder to reconstruct these manipulated latent codes to corresponding transformed images, i.e., $dec : Z \rightarrow X $. 
%Disentanglement means that one latent code reveals one semantic feature only. 
Disentanglement means that factorized attributes could be recognized and manipulated from encoded latent codes. 
\textit{Latent  Codes  Expansion -- LCE} is  applied  to  boost the image reconstruction quality. \textit{Unsupervised  Attribute  Directing  and  Manipulating -- ADM} is implemented to find latent directions towards specific attributes change via conducting principle analysis on the latent codes, and then obtain interpretable manipulation via adding bounded perturbation into the corresponding latent codes along the corresponding direction for generating nature transformations. Therefore, we can conduct various $T_s$ by reconstruction of the manipulated corresponding latent codes.

% %\vspace{-4mm}
\begin{figure*}[!htb]
	\centering
	\setlength{\abovecaptionskip}{-0.05cm}
	\setlength{\belowcaptionskip}{-0.2cm}
	\includegraphics[width=5.8in,height=2.5in]{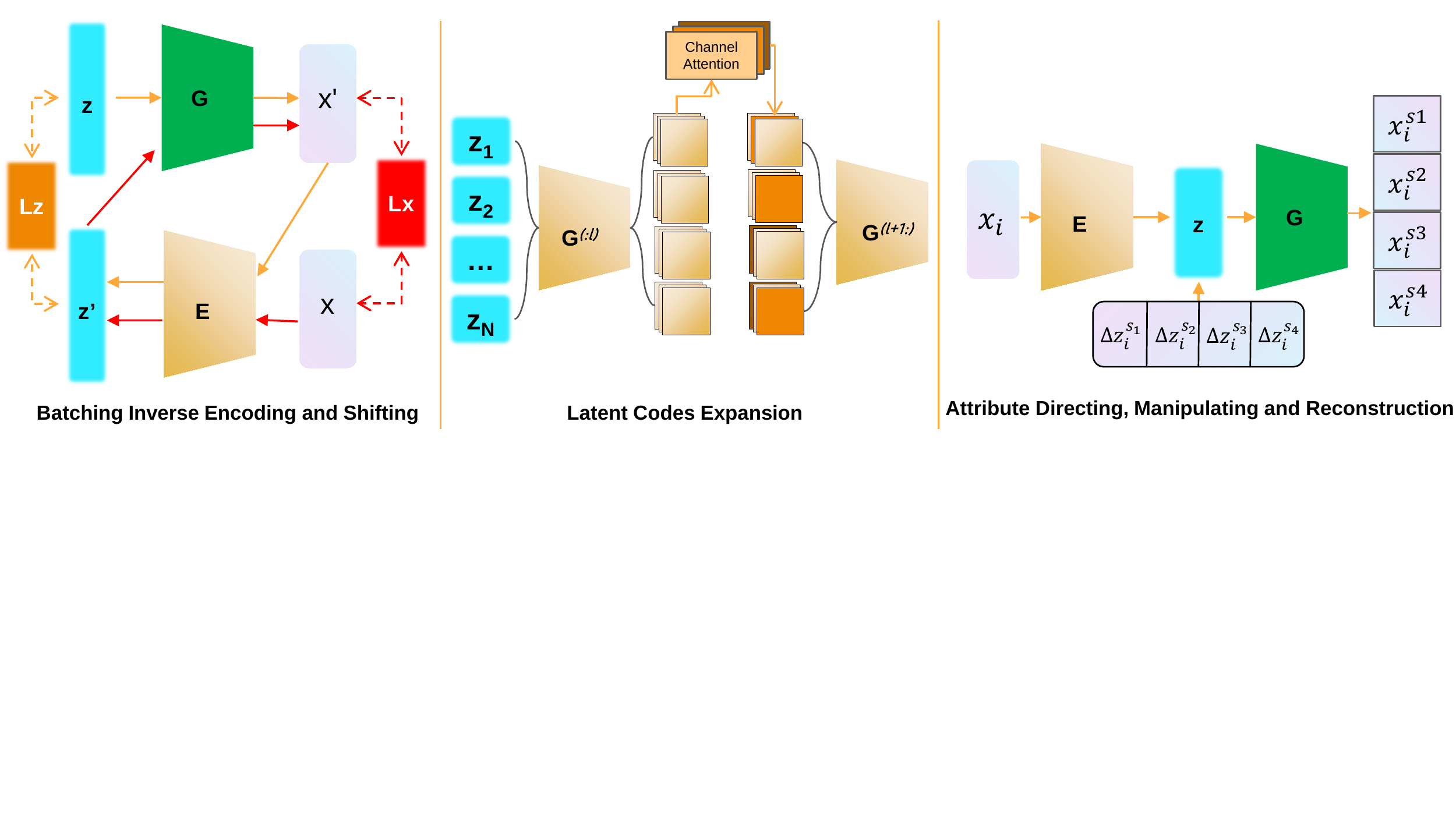}
	\caption{The scheme of natural transformation (NaTra).}
	\label{fig:3}
\end{figure*}
% %\vspace{-2mm}

Formally, for a C-class ($C\geq2$) classification problem, given a dataset $\{x_i, y_i\}_{i=1,\cdots,n}$ with $(x_i,y_i)$ is a pair of example and corresponding label from the data distribution $D \subset X \times Y$, $n$ is the the number of training examples, a DNN classifier $f_\theta$ parametrized by $\theta$, we aim to improve its robustness against variations derived from some natural transformations $T_{ s_{1:M}}:dec(z=enc(x)) \mapsto dec(\tilde{z}) $ in terms of a set of desired semantic features $s \in S$ via manipulating the corresponding latent codes after encoding $x_i$. Namely, we aim to refine the model parameters $\theta$ by solving the following min-max optimization problem:
\begin{equation}
    \underset{\theta}{min}\frac{1}{n}\sum_{i=1}^{n}\underset{\boldsymbol{\tiny{\begin{matrix}
T_s
\\ 
\tilde{z}_i^s \in B(\epsilon,z_i^s)
\end{matrix} }}}{max}L(f_\theta(T_s(x)),y),~\forall s \in S
    \label{eqi}
\end{equation}
% Given a deep classifier $f_\theta$ parametrized by $\theta$, we aim to improve its robustness against variantions derived from some transformations $T_{ s_1:s_M} $ in terms of a set of desired semantic features $s \in S$. Namely, we aim to refine the model parameters $\theta$ that minimize the loss caused by various variantions of transformations:
% \begin{equation}
%     \underset{(x,y)\sim D}{E}[\underset{T_s, s \in S}{max}L(f_\theta(T_s(x)),y)]
%     \label{eqi}
% \end{equation}
where $L$ can be any loss function (e.g. cross entropy loss or 0-1 loss in the context of classification tasks $\textbf{1}(f_{\theta}(T_s(x))\neq y)$). 
A specific transformation $T_s \in T_{ s_1:s_M}$ is expected to produce real-world natural transformation on a semantic feature $s$ from a set of desired semantic features $ S $, $\tilde{z_i} \in %B_{\epsilon}=\{\Delta_z:\left \| z_i-\tilde{z}_i \right \|_p \leq \epsilon\}$
B(\epsilon, z_i) := \{\tilde{z}_i:\left \| z_i-\tilde{z}_i \right \|_p \leq \epsilon\}$
denotes the $L_{p}$-norm ball centered at original latent codes $z_i$ with radius $\epsilon$.  For the robust classifier $f_\theta: X \rightarrow Y$, we would like that $f_\theta (T_s(x)) =f_\theta (x), \forall s \in S $. In particular, comprehensive task-irrelevant feature variations should be well investigated. For example, a face recognition classifier should not be affected by changes in the lighting conditions, background, or facial expression.

Based on the disentanglement, given a classification task, e.g., face recognition, that predicts the label $y$, $\tilde{z}_{\oslash}^s$ denotes the manipulated latent codes along the direction of a specific task-irrelevant semantic feature $s$, such as the facial expression.
Given an example and labelled pair $(x,y)$, we can obtain the disentangled latent codes via $z=enc(x)$, then for any task-irrelevant feature $s$, the prediction label will not be affected when modifying corresponding $z_{\oslash}^s$ part of $z$.  %$z_{\setminus \oslash}$ denotes the latent codes excluding $z_{\oslash}$.
%\begin{equation}
%%    P(y|dec(z_{\setminus \oslash}))=P(y|dec(z))
% P(y|dec(z \pm \epsilon \, dz_{\oslash}))=P(y|dec(z)).
% \end{equation}
%where $z_{\setminus \oslash}$ is the latent codes excluding $z_{\oslash}$.
The transformation $T_s$ on a specific semantic feature $s$ can be reflected in the perturbed latent code $\tilde{z}_{\oslash}^s$. Formally, given the desired semantic feature set S, the \textit{natural transformation} for a specific instance $x$ with label $y$ is produced via
\begin{equation}
\begin{aligned}
T= \{(T_s| T_s(x) = dec(\tilde{z}_{\oslash}^s),  \\
s.t.~ f(dec(z)) = f(dec(\tilde{z}_{\oslash}^s))\}, \\ \forall s \in S ~ and ~ \tilde{z}_{\oslash}^s \in B_{(\epsilon,z)}. 
\end{aligned}
\end{equation} 
%where $z_{\setminus \oslash},\tilde{z}_{\oslash}^s$ means we modify the task irrelevant part of latent codes along the direction in terms of a specific attribute $s$, i.e. $\tilde{z}_{\oslash}^s$, and keep the rest $z_{\setminus \oslash}$ unchanged.
Therefore, the robust classifier $f^*$ should predict a given $x$ with the correct label while resisting bounded variations on its latent codes and for all desired task-irrelevant features. 
%Namely, 
% \begin{equation}
% \begin{aligned}
%     f^*(dec(z))=f^*(dec(\tilde{z}_{\oslash}^s)),  
%     \\
%     \forall s \in S ~ and ~ \tilde{z}_{\oslash}^s \in B_{(\epsilon,z)}
% \\
% \underset{y\in Y}{argmax}f^*(dec(\tilde{z}_{\oslash}^s))=y
% \end{aligned}
% \label{eqf}
% \end{equation} 

A robust model $f_{\theta}^*$ to $ T $ means there is no bounded perturbation on any $T_s$ would cause the misclassification of $f_\theta^*$. 
Formally, the robustness depends on parameters $\theta^*$ that satisfies Equation \ref{eqtheta}. 
\begin{equation}
    \theta^*=\underset{\theta}{argmax}\underset{(x,y)\in D}{E}[\underset{s\in S, \tilde{z}_{\oslash}^s \in B_{(\epsilon,z)}}{max}L(f_\theta(dec(\tilde{z}_{\oslash}^s)),y)]
\label{eqtheta}
\end{equation}
Solving Equation \ref{eqtheta} requires solving the corresponding inner-maximization problem for each $ z_{\oslash}^s$ within bounded ball:
\begin{equation}
    \tilde{z}_{\oslash}^*=\underset{s\in S, \tilde{z}_{\oslash}^s \in B_{(\epsilon,z)}}{argmax}L(f_\theta(dec(\tilde{z}_{\oslash}^s)),y)
\end{equation}

%As enumerating all possible latents $ \tilde{z}_n\in Z_n$ is often intractable, we resort to a technique popularized by Madry et al. [19] in the context of adversarial training, which consists of using projected gradient ascent on a differentiable surrogate loss. 
$L$ can be replaced with the cross-entropy loss $\hat{L}$: 
\begin{equation}
    \hat{L}(f_{\theta}(x),y)=-log([f_{\theta}(x)]_y)
\end{equation}
where $[a]_i $ returns the $i^{th}$ coordinate of a. Gradient ascent steps are then interleaved with projection steps for a given number of iterations $K$. $\tilde{z}_{\oslash}^*$ can be estimated by recursion interleaved with projection steps for $K$ iterations for each $s$, i.e. $ \tilde{z}_{\oslash}^{(K)}$: 
\begin{equation}
    \tilde{z}_{\oslash}^{(k+1)}=proj_{Z_{\oslash}}(\tilde{z}_{\oslash}^{(k)}+\alpha \triangledown_{\tilde{z}_n^{(k)}}\hat{L}(f_{\theta}(dec(\tilde{z}_{\oslash}^{(k)})),y))
\end{equation}
where $\alpha$ is a constant step-size and $proj_A(a)$ is a projection operator that project $a$ onto $A$ as \cite{gowal2020achieving}.

The baseline approach to achieve the robustness is to fine-tune the model $f$ on the additional data that consists of a set of natural transformations $NT_x^s, \forall s \in S$ with bounded perturbation for each training instance with correct label. 
After finding the latent direction for a specific task-irrelevant semantic feature, we can construct the natural transformation set in terms of the features via editing the latent codes via the decided direction. 
Consequently, the natural transformation augmentation is defined as:
\begin{equation}
\begin{aligned}
NT_x = \{(T_s(x),y) | T_s(x) = dec(\tilde{z}_{\oslash}^s),  \\
s.t.~ f(dec(z)) \neq  f(dec(\tilde{z}_{\oslash}^s)\}), \\ \forall s \in S ~ and~ \tilde{z}_{\oslash}^s \in B_{(\epsilon,z)}. 
\end{aligned}
\end{equation} 

\subsection{Batch  Inverse  Encoding and Shifting}
By learning to map into the latent space of a pre-trained image generator, we leverage both its state-of-the-art generative power, and its disentangled and expressive latent space, without the burden of training it. We train an additional encoder to obtain the inverse mapping from massive given images to the latent space of a well-trained generator network. Such an encoder provides a fast solution of image embedding by performing a forward pass through the encoder neural network.
%Cyclic parameter optimization is used for training this inverse mapping. 

Given a generator network $g_{\nu}(z)$ (i.e. $dec_{\nu}(z)$) that maps the latent space $Z$ to the data space $X$, BIES aims to learn an encoder network $enc_{\eta}(x)$ that maps the image $x$ from the data space to the latent space $Z$ of $g_{\nu}(z)$. 
The architecture of BIES is shown at the left of Figure~\ref{fig:3}.
The first goal of BIES is the ability to reverse the latent representation of randomly generated samples. For a given random $z$, the generator generates an image $x'$ that is then fed into the encoder network to obtain an estimated  latent codes $z'$. We minimize the objective in Equation $\ref{eqz}$ to minimize the error between the estimated $z'=enc_{\eta}(g_{\nu}(z))$, and the initial $z$. 
\begin{equation}
    \eta^*= \underset{\eta}{argmin}(E_{z\sim P(Z)}\left \| z -enc_{\eta}(g_{\nu}(z)) \right \|^2_2) 
\label{eqz}
\end{equation}

The second goal of BIES from the model is to reconstruct real images from $z'$ with good quality. For a given image $x$ and its encoded $z'$, the generator $g$ is used to reconstruct its corresponding image $x'=g_{\nu}(enc_{\eta}(x))$. We use the objective in Equation $\ref{eqx}$ to minimize the error between $x'$ and $x$. To improve the reconstruction quality, both pixel-wise MAE error (low-level) and perceptual error (high-level) are used to better steer the optimization. The perceptual loss is evaluated on the the $l_1$ distance between the perceptual features $V_l(.)$ extracted at the $i^{th}$ layer of a trained VGG-16 network. 
\begin{equation}
\begin{aligned}
        \eta^*=\underset{\eta}{argmin}(E_{x\sim P_{data}} (\gamma_1 \left \| x-g_{\nu}(enc_{\eta}(x)) \right \|_2^2 +\\ \gamma_2 \sum_{l} \left \| V_l (x)- V_l (g_{\nu}(enc_{\eta}(x)) \right \|_1))
\end{aligned}
\label{eqx}
\end{equation}
 
The encoder's parameters are updated twice in each iteration in terms of  Equation $\ref{eqx}$ and $\ref{eqz}$, respectively.  The input of the encoder is alternatively either real or generated images, and the input of the generator is alternatively either a randomly generated or an encoded latent vector as the input. During training, only the parameters $\eta$ are updated,  while the well-trained generator is fixed. 
The encoder is trained using a pre-trained generator without facing stability problems. 

After the encoder is well trained, the generator will be fine-tuned using the instances in the target domains, to shift the weights of the generator as a decoder towards the target distribution.
We fix the encoder while updating the weights of the generator to match the targeted generated image. 
The pre-trained generator will be shifted toward the targeted domain after slightly updating its weights on the targeted domain images. 
For each image, we fix its encoded latent codes and update the weights of $g$, dragging the nearest neighbor of the pre-trained generator manifold closer to the target manifold. 
The objective function can be defined as: 
\begin{equation}
\begin{aligned}
   g^*=\underset{\nu}{argmin}(E_{x\sim P_{data}} (\gamma_1 \left \| x-g_{\nu}(enc_{\eta}(x)) \right \|_2^2 + \\\gamma_2 \sum_{l} \left \| V_l (x)- V_l (g_{\nu}(enc_{\eta}(x)) \right \|_1))
\end{aligned}
\label{eqx}
\end{equation}
Equation $\ref{eqx}$ is used as the loss function for fine-tuning $g$. 

%To further improve efficiency, layer-wise fine-tuning is adopted instead of the entire generator. Front layers of generator always control the global attributes such as gender, appearance, or identity. In contrast, the late layers control the lowlevel details of the output image, such as the color or local textures \cite{ONE,23}. 
%To better illustrate the effects of manifold projection and shifting, \llv{where is the figure???}Fig. 2 shows visual examples of an input, the reconstructed image after BIES, and the reconstructed image after manifold shifting. It shows that after manifold shifting, the reconstructed image matches with the input more closely in global color and appearances.
%Training details: 
%All of the weights of the convolutional layers of the encoder, including the pre-trained weights of the VGG Network, are trained using the same learning rate. We train the encoder using RMSProp optimizer [38] with a batch size of 128, setting rho to 0.9 and epsilon to 1e-08. To adapt the existing VGG layer weight values properly, we use a small learning rate, i.e. 1e-4, for optimizing all the parameters in the encoder architecture. We reduce the learning rate by 2 when minimum validation loss stops improving for ten epochs. After the training process, we expect that the generated images and their reconstructions look almost the same and the reconstructions of the real images are at an acceptable level; sufficient to find accurate latent vector directions for the attribute embedding. 
\subsection{Latent Codes Expansion}
Given a target image $x$, the BIES can be applied to reverse the generation process by encoding the latent code to recover $x$. However, it is hard to reconstruct an ideal image by optimizing a single latent code vector, which may not be enough to recover all the details of a particular image. 
In this strategy, we embed given images into an extended latent space, which is a concatenation of multiple latent vectors, used for revealing multiple feature maps and improving image reconstruction quality \cite{gu2020image}. These extended latent vectors will be composed at some intermediate layer of the generator, in terms of adaptive channel importance to better recover the input image. 
Namely, latent codes $z$ are extended to $z^+$, and they are incorporated together via merging corresponding intermediate feature maps with the attention mechanism. $N$ latent codes $\{z^i\}_{i=1}^{N}$ are implemented to reverse given inputs, each of which can help reconstruct some sub-regions of the target image. 
In particular, a specific layer (with index $l$) of generator $g(·)$ is selected to divide $g$ into two sub-networks, i.e., $g^{l}_h(·)$ and $g^{l}_r(·)$. For any $z^i$, we can extract the corresponding spatial feature $gc^{l}(·)=\sum_i^N g^{l}_h(z_i)$ for further composition. 
Each $z^i$ is expected to recover some particular regions of the target image. It has been demonstrated that different channels of the generator control different visual concepts such as objects and textures \cite{bau2018gan}. Consequently, the attention mechanism is used to allocate channel importance $\alpha_i$ for each $z^i$ to encourage them align with different semantics. Here, $\alpha_i \in R^C$ is a $C$-dimensional vector to reveal the importance of the corresponding channel of the feature map, and $C$ is the number of channels in the $l$-th layer of $g(·)$.  The reconstructed image can be generated with $x'= g_r(\sum_{i=1}^{N} gc^{l} \odot \alpha_i)$ where $ \odot $ denotes the channel-wise multiplication as $\{gc^{l} \odot \alpha_i\}_{h,w,c}= \{gc^{l}\}_{h,w,c} \times \{\alpha_i\}_c$.
Here, $h$ and $w$ indicate the spatial location, while $c$ stands for the channel index.

StyleGANs have two kind of latent space, i.e. initial latent space $Z$ and the intermediate latent space $W$. The 512-dimensional vectors $w \in W$ are mapped from the 512-dimensional vectors $z \in Z$ by a fully connected neural network $M$. The extended latent space $W^+$ is a concatenation of 18 different 512-dimensional $w$ vectors, one for each layer of the StyleGAN architecture that can receive input via adaptive instance normalization (AdaIN) \cite{huang2017arbitrary}.

\subsection{Unsupervised Attribute Directing, Manipulating and Reconstruction (ADMR)}
To identify latent directions corresponding to important semantic features, a dimensionality reduction approach based on uniform manifold approximation and projection (UMAP) is implemented in the activation space. Layer-wise manipulation of the GANs are then performed to produce edits in the input image that are interpretable in terms of chosen semantic features \cite{PCA}. 
ADMR has two advantages: it is simple and unsupervised. 
\subsubsection{Attribute Directing} 
The first step is to discover valuable directions in the latent space for some semantic features. 
Generally, the GAN network consists of a sequence of layers $g^{1}, \cdots, g^{L}$. The first layer adopts the latent vector as input and provides a set of activations $ o_1 = g^{(1)}(z)$. The remaining layers each produce activations as a function of the previous layer's output.  The last layer's output $x' = g^{(L)}( o_{L-1})$ is an image. 
For StyleGAN \cite{karras2019style,karras2020analyzing}, the first layer takes a constant input, $ o_0$. The output is controlled by a non-linear function of $z$ as input to intermediate layers, $ o_i = g^{(i)}( o_{i-1},w=M(z))$  where $M$ is an 8-layer multilayer perceptron. 
For the BigGAN \cite{brock2018large}, the intermediate layers also use the latent vector as input, $ o_i = g^{(i)}( o_{i-1},z)$.

The principal components of activation tensors on the first few layers of the generator represent important factors of variation (\cite{PCA}). ADMR aims to recognize the principal axes of $p(w)$ or $p(z)$. 
Generally, $K$ random latent codes tensors are sampled, i.e. $z^{D*N}_{\{1:K\}}$ where $D$ is the dimension of each latent codes (e.g. 512), $N$ is the extension scale for Latent Codes Expansion, and $K$ is the number of sampling. The dimensionality reduction algorithm UMAP is used to calculate the principal components of these latent codes, resulting in a low-rank basis $V$.  

For StyleGAN, we then figure out the corresponding $w^i = M(z^i)$ values. We can edit the image, with the encoded latent codes $w$, via $w'=w+Vr, x'=g(o_0,w')$ where each entry $r_i$ of $r$ is a separate control parameter. 

For BigGAN, we first perform UMAP at on the intermediate network layer $i$ of the generator. 
Namely, $N$ random latent codes are sampled to create $N$ activation output vectors at the $i^{th}$ layer. We then conduct UMAP on these $N$ activation vectors to provide a low-rank basis matrix $V$, and the data mean $\mu_U$. The UMAP coordinates $b_j$ of each activation are then computed by projection: $b_j=V^T(o_j-\mu_U)$.
Then we transfer these directions to the latent space $z$ by linear regression. Given an individual basis vector $v_k$ (i.e., a column of $V$), and the corresponding coordinates $b^i_{1:N}$, where $b^i_{j}$ is the scalar $i^{th}$ coordinate of $b_j$. The latent basis vector $u_k$ to identify a latent direction corresponding to this principal component is given as follows: 
$u_i=arg ~min \sum_j \left \| u_ik_j^i-z_j \right \|^2$.
Equivalently, the whole basis is computed simultaneously with: 
\begin{equation}
U=arg~ min \sum_j \left \| Uk_j-z_j \right \|^2
\end{equation}
 
Each column of U then aligns with the variation along the corresponding column of V. The individual dimensions $b_i$ each correspond to different edits in an interpretable manner. Given a new image with latent codes $z$, manipulation can be conducted by varying the coordinates of $x$: 
\begin{equation}
z'=z+Ub,x'=g(z')
\end{equation}
Besides StylecGAN and BigGAN, experiments show that ALM yielded useful directions for earlier models, including DCGAN and Progressive GAN, using the BigGAN procedure above. 
 
%We use incremental PCA [18] for efficient computation, and use $N = 10^6$ samples. On a relatively high-end desktop PC, computation takes 2 minutes on StyleGAN and StyleGAN2. Our results include edits discovered from StyleGAN (Bedrooms, Landscapes, WikiArt training sets), and StyleGAN2 (FFHQ, Cars, Cats, Church, Horse training sets). Our StyleGAN model weights were obtained from https://github.com/justinpinkney/awesome-pretrained-stylegan, except for Landscapes, which was provided by artbreeder.com. Our StyleGAN2 models were those provided by the authors online [13]. 

\subsubsection{Attribute Manipulating}
Given the directions found with UMAP, we now show that these can be decomposed into interpretable edits

For StyleGAN, layer-wise control of StyleGAN is conducted on the latent codes $w^i$. 
We use the notation $D(u_i, j-k)$ to denote edit directions; for example, $D(u_1, 0-3)$ means moving along component $u_1$ at the first four layers only. $D(u_2, all)$ means moving along component $u_2$ globally: in the latent space and all layer inputs. 
%For example, component $u_1$ that controls head rotation and gender in an entangled manner when applied to all layers, can be forced into a much more pure rotation when only applied to the first three layers in $D(u_1, 0-2)$; similarly, the age and hairstyle changes associated with component u4 can be removed to yield a cleaner change of lighting by restricting the effect to later layers in $DE(u_4, 5-17)$. It is generally easy to discover surprisingly targeted changes from the later principal components. Examples include $D(u_10, 7-8)$ that controls hair color, as well as $E(u_11, 0-4)$ that controls the height of the hair above the forehead. More examples across several models are shown in Figure 5. As shown in Figure 6, multiple edits applied simultaneously across multiple principal directions and internal layer ranges compose well. We find that some of these learned edits are dependent on the identity of the starting face. For example, on StyleGAN2 faces, a "beard" edit only modifies male faces, and the "lipstick" edit only modifies female faces, reflecting biases learned from the training data. Many of the edits are not fully disentangled, varying several image attributes at once. 
A simple user interface can be provided for the exploration of the principal directions in an interactive manner. After computing the directions, the user is free to inspect the effect of any of them by simple slider controls. The layer-wise application is enabled by specifying a start and end layer for which the edits are to be applied. 
%The GUI also enables the user to name the discovered directions, as well as load and save the results. 
%The exploration process is demonstrated in the video. \llv{where is the video???} The sliders operate in units of standard deviations, and we find that later components work for wider ranges of values than earlier ones. 
The first ten or so principal components, such as head rotation $(D(u_1, 0-2))$ and lightness/background $(D(u_8, 5))$, operate well in the range $[-2, \cdots, 2]$, beyond which the image becomes unrealistic. In contrast, face roundness $(D(u_{37}, 0-4))$ can work well in the range $[-20, \cdots, 20]$, when using 0.7 as the truncation parameter. For truncation, we use interpolation to the mean as in StyleGAN. The variation in slider ranges described above suggests that truncation by restricting $w$ to lie within two standard deviations of the mean would be a very conservative limitation on the interface's expressivity since it can produce interesting images outside this range. 
% There are several benefits of using PCA directions. First, PCA identifies and orders the most significant directions, such as global rigid motion.
% Moreover, we find that the PCA directions decompose into interpretable edits more naturally. Importantly, PCA places all object motion and camera motion (e.g., head rotation) into the first ten or so components, and all later components have motion factored out. In contrast, nearly all random directions can be expected to include some motion.

For other GANs that do not support layer-wise edits, e.g., BigGAN, they can be modified to produce behavior similar to StyleGAN.  This is conducted by varying the intermediate Skip-z inputs $z_i$ separately from the latent $z$. Namely, $z_i$ is enabled to vary individually between layers in a direct analogy to the style mixing of StyleGAN. Edits may be performed to the inputs to different layers independently.

\subsubsection{Adaptive Perturbation Tuning for Construction of Nature Transformation}
To prevent the edited latent codes from going far from the manifold, \emph{Adaptive Perturbation Tuning} (APT) is used to bound the manipulation of the latent codes.

Perturbation noise can be obtained by randomly sampling from $N(0,1)$ for $N_r$ times and projecting perturbed latent code $\tilde{z}_{\oslash}^{(K)}$ back onto a $l_\infty$-bounded neighborhood around $\tilde{z}_n^{(0)}: \{z_n| \left \| \tilde{z}_n^{(0)} – z_n \right \|< \epsilon \}$ where $\epsilon$ is set to 0.03.

As a fixed constant, the perturbation threshold $\epsilon$ ignores the fact that every data example may have different intrinsic robustness \cite{cat}. To address the accuracy-robustness trade-off, adaptive threshold tuning is used to discover a non-uniform and effective perturbation level and the corresponding customized target label for each example. Intuitively, for training samples that are intrinsically closer to the decision boundary, a smaller $\epsilon$ will be applied to reduce the generalization error introduced by adversarial training. 

The perturbation level $\epsilon_i$ allocated to each example $x_i$ is given as 
\begin{equation}
   \epsilon_i = \underset{\epsilon}{argmin} \{ \underset{\tilde{z}_i \in B(\epsilon,z_i)}{max} f_{\theta} \neq y_i\}
    \label{eqadp}
\end{equation}

Alternative updates are used to update each $ \epsilon_i $, i.e., implementing one SGD update on $\theta$, and then updating the $ \epsilon_i $ in the current batch. Specifically, at each iteration, we conduct the nature transformation with perturbation sensitivity $\epsilon_i +\rho$ where $\rho$ is a constant. If the misclassification occurs, then we keep the current $ \epsilon_i $. If not, we increase $\epsilon_i +\rho$ until misclassification occurs or $\epsilon_i$  reaches the truncated value to make sure $\epsilon_i$ will not be too large. 
As the transformations are also used to update the model parameter $\theta$, no additional cost is introduced by the adaptive threshold tuning. 

Besides, the adaptive label smoothing strategy is also used to reflect different perturbation tolerance on each example. Label smoothing is proposed in \cite{szegedy2016rethinking} to convert one-hot label vectors into one-warm vectors $\title{y}=(1-\alpha)y + \alpha u$ representing low-confidence classification to prevent the model from making over-confident predictions. $\alpha$ is set as $c\times \epsilon_i$ so that a more significant perturbation sensitivity would hold a higher label uncertainty, and c is a hyperparameter. $u$ is set as $Dirichlet(1)$, namely Dirichlet distribution on an all one vector \cite{cat}. 

Consequently, the objective function in Equation \ref{eqi} is updated as 
\begin{equation}
    \begin{aligned}
         \underset{\theta}{min}\frac{1}{n}\sum_{i=1}^{n}\underset{\boldsymbol{\tiny{\begin{matrix}
T_s, s \in S
\\ 
\tilde{z}_i^s \in B(\epsilon_i,z_i^s)
\end{matrix} }}}{max}L(f_\theta(T_s(x)),\tilde{y})
    \end{aligned}
\end{equation}

%Figure 1 shows the result of this optimization procedure where the original image (on the top left) is classified as “not smiling,” and the optimized image (on the bottom-left) is classified as “smiling.” Once latents $\tilde{z}_{\oslash}= \tilde{z}_{\oslash}^{(K)}$ are found, we can compute the cross-entropy loss on the generated natural transformation image to update $f_{\theta}$ by solving Equation $\ref{eqtheta}$. 
\iffalse
\begin{equation}
\begin{aligned}
    \underset{\theta}{argmin}E_{z_{i,c},y_i\in D}[L(f_{\theta}(dec(z_{i,c},\tilde{z}_n)), y_i)]\\
and~\tilde{z}_n=\underset{z_n \in Z_n}{argmax}L(f_{\theta}(dec(z_{i,c},z_n)), y_i)
\end{aligned}
\end{equation}
\fi

%Random mixing with disentangled representations. While this section describes an instantiation of AdvMix using StyleGAN, it is possible to formulate an equivalent random data augmentation baseline. For an input x, we generate a random variation as follows: 
% \begin{equation}
%     \tilde{x}=dec(enc(x)_c, map(z)_n) with z \sim N(0, 1)
% \end{equation}
\section{Experiments}
\subsection{Data and Setup}
%\bf{Surya - put some high resolution example image results in Appendix; we can have 5 pages references and Appendix; we can use that space to show that we have done an extensive experiments and some results} 
The performance of NaTra is evaluated on three datasets: LSUN car (resized to 512 $\times$ 512) and cat (256 $\times$ 256) \cite{yu2015lsun}, and CelebA-HQ  \cite{karras2019style,karras2020analyzing} (Face). 

%Color-MNIST consists of a dataset of MNIST [55] digits that are artificially colored to emphasize bias. On the training set, we color each pair (x, y) of the original MNIST dataset with a color drawn randomly from a normal distribution with mean $\mu_y$ and standard deviation $\sigma$ (means $\mu_y$ for $y \in \{0, \cdots, 9\}$). On the test set, we color digits uniformly at random. In other words, the colors present in the training set spuriously correlate with the label. We can use $\sigma$ to affect this correlation: by progressively increasing $\sigma$ the dataset becomes less biased. For all techniques (including mixup), we vary the level of bias and train models using 5 epochs. The StyleGAN model is trained on the training set only, once for each setting of $\sigma$. The disentangled latents defining the finer style correspond to the final resolution of 32 $\times$ 32.
%CelebA-HQ is a large-scale public dataset with forty different face attribute annotations including whether a person smiles or wears a hat.
We make no modifications to the dataset and use a pre-trained StyleGAN model. For all techniques, we train models using 20 epochs. We evaluate all methods on their ability to classify the smiling attribute, as well as three other attributes. In this experiment, the disentangled latents defining the finer style correspond to resolutions ranging from 128 $\times$ 128 to 1024 $\times$ 1024. CelebA-HQ contains 30,000 high quality images at 1024 $\times$ 1024 resolution. 

We use both standard VGG-16 \cite{simonyan2014very} and WideResNet for robustness performance evaluation. For VGG-16, we implement the standard hyperparameters. For WideResNet, we use the same model structure provided by \cite{madry2017towards}. 
%In all datasets, 20,000 clean examples are randomly selected for training NaTra. We randomly select 5,000 clean images (named CLE, labeled 0) and generating 5,000 successfully attacked adversarial examples (named ADV, labeled 1), respectively. These datasets are used to test the efficiency of the VASA-Defense. A further 2,000 clean instances are randomly chosen as the validation data (named VAL) to decide thresholds or train the detector discriminator. 
We use four existing data augmentation approaches as baselines compared with NaTra: 
(1) Random perturbation (Random). For an input $x$, we generate a random variation via adding random noise from $N(0,1)$ to the latent codes; (2) Adversarial Training (Adv) which minimizes the adversarial risk over $l_\infty$-norm bounded perturbations of size $\epsilon_{pixel}$ in input space; (3) Mixup in the latent space (MixL). It augments data by reconstructing mixed latent codes between the two pairs of inputs; and (4) Mixup in the input space (MixI). It augments data by mixing two pairs of inputs in the input space; (5) NaTra without Latent Codes Expansion (NaTra-OL); (6) NaTra without Adaptive Perturbation Tuning (NaTra-OA); (7) AdvMix \cite{advmix}.
\subsection{Robustness Evaluation and Analysis}
In this section, we evaluate the robustness of NaTra on benchmark datasets. Finally, we benchmark the state-of-the-art robustness and %explore using 
integrate the unlabeled data for further improvement.

% Please add the following required packages to your document preamble:
% \usepackage{multirow}
% %\vspace{-2mm}
\begin{table}[!htb]
\centering
\caption{Test accuracy on attributes (i.e. with glass, gender, smiling, age and identification, denoted by S1-S4 and ID) classification tasks on Face dataset with different robust enhancements.}
\label{tab:my-table}
\begin{tabular}{c|lllll}
\hline
\multirow{2}{*}{Method} & \multicolumn{5}{l}{Accuracy on Attributes (\%)} \\ \cline{2-6} 
                        & S1       & S2      & S3      & S4      & ID      \\ \hline
Original                & 96.82    & 90.32   & 84.16   & 79.76   & 75.87   \\
Adv                     & 95.86    & 91.41   & 84.28   & 79.81   & 75.69   \\
MixI                    & 95.92    & 90.35   & 84.18   & 79.96   & 76.12   \\
MixL                    & 96.91    & 90.41   & 84.28   & 79.81   & 75.69   \\
Random                  & 97.12    & 91.56   & 86.12   & 81.79   & 76.16   \\
NaTra-OL                & 97.56    & 92.29   & 85.65   & 79.47   & 76.21   \\
NaTra-OA                & 97.83    & 92.35   & 85.72   & 79.55   & 76.41   \\
NaTra                   & 98.11    & 93.10   & 86.33   & 80.18   & 77.02   \\ \hline
\end{tabular}
\end{table}
% %\vspace{-3mm}

% %\vspace{-2mm}
\begin{table}[!htb]
\centering
\caption{Test accuracy on classification tasks on LSUN Car and Cat datasets with different robust enhancements.}
\label{tab:my-table2}
% \begin{tabular}{c|llllllll}
% \hline
% Accuracy (\%) &
%   \multicolumn{1}{c}{Original} &
%   \multicolumn{1}{c}{Adv} &
%   \multicolumn{1}{c}{MixI} &
%   \multicolumn{1}{c}{MixL} &
%   \multicolumn{1}{c}{Random} &
%   \multicolumn{1}{c}{NaTra-OL} &
%   \multicolumn{1}{c}{NaTra-OA} &
%   \multicolumn{1}{c}{NaTra} \\ \hline
% Cat &
%   92.36 &
%   92.50 &
%   92.37 &
%   92.33 &
%   93.25 &
%   93.03 &
%   93.62 &
%   94.15 \\
% Car &
%   90.10 &
%   91.05 &
%   89.80 &
%   89.83 &
%   90.69 &
%   91.76 &
%   92.14 &
%   92.86 \\ \hline
% \end{tabular}
\begin{tabular}{c|ll}
\hline
\multirow{2}{*}{Method} & \multicolumn{2}{l}{Accuracy (\%)} \\ \cline{2-3} 
                        & Cat                       & Car                      \\ \hline
Original                & 92.36                     & 90.10                    \\
Adv                     & 92.50                     & 91.05                    \\
MixI                    & 92.37                     & 89.80                    \\
MixL                    & 92.33                     & 89.83                    \\
Random                  & 93.25                     & 90.69                    \\
NaTra-OL                & 93.03                     & 91.76                    \\
NaTra-OA                & 93.62                     & 92.14                    \\
NaTra                   & 94.15                     & 92.86                    \\ \hline
\end{tabular}
\end{table}
%\vspace{-4mm}
\begin{figure}[!htb]
	\centering
	\setlength{\abovecaptionskip}{-0.05cm}
	\setlength{\belowcaptionskip}{-0.2cm}
	\includegraphics[width=3.5in,height=3.0in]{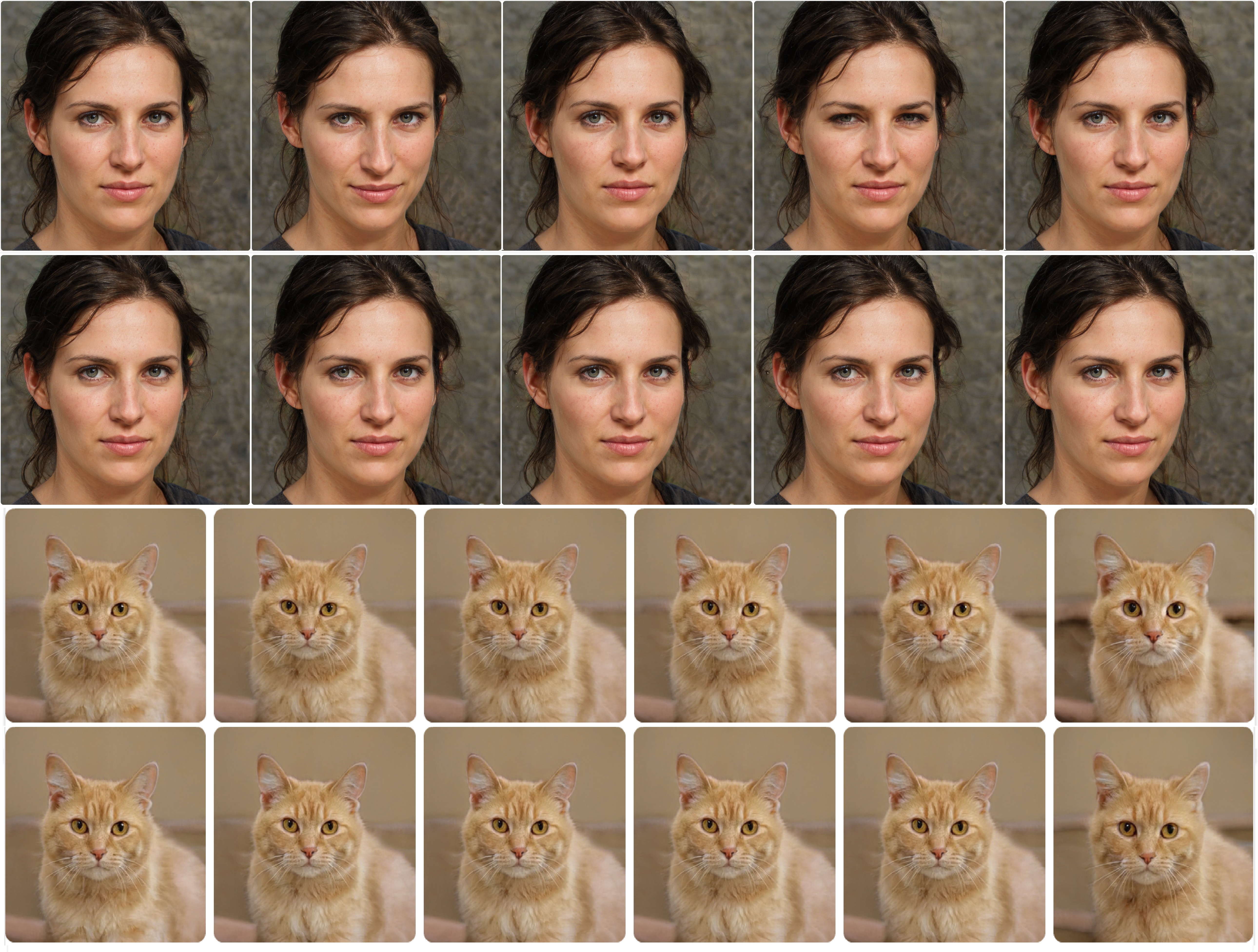}
	\caption{The demonstration of escaped samples that are misclassified by the actual model, and correctly classified by fine-tuning the model via NaTra on natural transformation, i.e., face orientation, eye-opening, facial expression for human, and head elevation and body orientation for cat.}
	\label{fig:4}
\end{figure}
%\vspace{-4mm}
The robustness evaluations of all data augmentation approaches are shown in Table~\ref{tab:my-table} with respect to the same classification task, where “Original” denotes the accuracy on natural test images. Our proposed NaTra achieves the best robustness on both CelebA and LSUN. Particularly, NaTra improves $\sim 8\%$ over Adv, and $\sim 4\%$ even over Mixup. A similar trend of improvement is also observed among NaTra and NaTra-OL and NaTra-OA, which reveals the efficiency of the LCE and Adaptive Perturbation Tuning. 

Compared with images of low fidelity, the robustness improvements of NaTra over other baselines are more significant on images of high fidelity. This is because adversarial training on high fidelity is a more challenging problem that may have more misclassified examples during training. 

We also evaluate the robustness evaluations of all data augmentation approaches with respect to various classification task in Table~\ref{tab:my-table2}. It is also confirmed that NaTra systematically achieves high accuracy, namely NaTra can lead to a lower generalization error.  

We find there are a large number of escaped examples of images that are all correctly classified by the actual model, but its transformations are misclassified. After fine-tuning using NaTra, these escaped examples can be correctly classified, as demonstrated in Figure~\ref{fig:4}.

It is often challenging, if not completely impossible, to collect a large-scale dataset for a certain person, an object, which is essential to train good deep models, e.g., classifiers. To address data limitation, few-shot learning in image generation \cite{zhao2020differentiable} or fine-tuning to transfer the knowledge of pre-trained models \cite{wang2018transferring} can be applied to train good generator and then implement NaTra. It is also interesting to find that it is feasible to train a face recognization model with more than $95\%$ accuracy when giving only 100 face images for each person. 

\section{Conclusion}
In this paper, we propose NaTra, an adversarial training scheme that is designed to improve the robustness of deep models against customized input variations arising in real world natural transformations. Our framework can be realized by harnessing pre-trained generative models to conduct nature transformations for data augmentation
%via Batch Encoding Reversal and Shifting, Latent Codes Expansion, and Unsupervised Attribute Directing, Manipulating and Reconstruction
, resulting in controllable attribute edit as well as good reconstruction quality. Experimental results demonstrated that NaTra could relieve the insensitivity to task-irrelevant variations while increasing the deep models' generalization. We hope this work could %contribute to the advancement of 
advance the implementation of robust deep models %towards safely implementing 
in the real world. 

{\small
\bibliographystyle{ieee_fullname}
\bibliography{ref}

\begin{thebibliography}{10}\itemsep=-1pt

\bibitem{abdal2019image2stylegan}
Rameen Abdal, Yipeng Qin, and Peter Wonka.
\newblock Image2stylegan: How to embed images into the stylegan latent space?
\newblock In {\em Proceedings of the IEEE international conference on computer
  vision}, pages 4432--4441, 2019.

\bibitem{baluja2017adversarial}
Shumeet Baluja and Ian Fischer.
\newblock Adversarial transformation networks: Learning to generate adversarial
  examples.
\newblock {\em arXiv preprint arXiv:1703.09387}, 2017.

\bibitem{bau2018gan}
David Bau, Jun-Yan Zhu, Hendrik Strobelt, Bolei Zhou, Joshua~B Tenenbaum,
  William~T Freeman, and Antonio Torralba.
\newblock Gan dissection: Visualizing and understanding generative adversarial
  networks.
\newblock {\em arXiv preprint arXiv:1811.10597}, 2018.

\bibitem{bau2019inverting}
David Bau, Jun-Yan Zhu, Jonas Wulff, William Peebles, Hendrik Strobelt, Bolei
  Zhou, and Antonio Torralba.
\newblock Inverting layers of a large generator.
\newblock In {\em ICLR Workshop}, volume~2, page~4, 2019.

\bibitem{brock2018large}
Andrew Brock, Jeff Donahue, and Karen Simonyan.
\newblock Large scale {GAN} training for high fidelity natural image synthesis.
\newblock In {\em International Conference on Learning Representations}, 2019.

\bibitem{carlini2017towards}
Nicholas Carlini and David Wagner.
\newblock Towards evaluating the robustness of neural networks.
\newblock In {\em 2017 ieee symposium on security and privacy (sp)}, pages
  39--57. IEEE, 2017.

\bibitem{cat}
Minhao Cheng, Qi Lei, Pin-Yu Chen, Inderjit Dhillon, and Cho-Jui Hsieh.
\newblock Cat: Customized adversarial training for improved robustness.
\newblock {\em arXiv preprint arXiv:2002.06789}, 2020.

\bibitem{creswell2018inverting}
Antonia Creswell and Anil~Anthony Bharath.
\newblock Inverting the generator of a generative adversarial network.
\newblock {\em IEEE transactions on neural networks and learning systems},
  30(7):1967--1974, 2018.

\bibitem{cubuk2018autoaugment}
Ekin~D Cubuk, Barret Zoph, Dandelion Mane, Vijay Vasudevan, and Quoc~V Le.
\newblock Autoaugment: Learning augmentation policies from data.
\newblock {\em arXiv preprint arXiv:1805.09501}, 2018.

\bibitem{engstrom2017rotation}
Logan Engstrom, Dimitris Tsipras, Ludwig Schmidt, and Aleksander Madry.
\newblock A rotation and a translation suffice: Fooling cnns with simple
  transformations.
\newblock {\em arXiv preprint arXiv:1712.02779}, 1(2):3, 2017.

\bibitem{geirhos2018generalisation}
Robert Geirhos, Carlos~RM Temme, Jonas Rauber, Heiko~H Sch{\"u}tt, Matthias
  Bethge, and Felix~A Wichmann.
\newblock Generalisation in humans and deep neural networks.
\newblock In {\em Advances in neural information processing systems}, pages
  7538--7550, 2018.

\bibitem{goodfellow2014explaining}
Ian~J Goodfellow, Jonathon Shlens, and Christian Szegedy.
\newblock Explaining and harnessing adversarial examples.
\newblock {\em arXiv preprint arXiv:1412.6572}, 2014.

\bibitem{gowal2020achieving}
Sven Gowal, Chongli Qin, Po-Sen Huang, Taylan Cemgil, Krishnamurthy Dvijotham,
  Timothy Mann, and Pushmeet Kohli.
\newblock Achieving robustness in the wild via adversarial mixing with
  disentangled representations.
\newblock In {\em Proceedings of the IEEE/CVF Conference on Computer Vision and
  Pattern Recognition}, pages 1211--1220, 2020.

\bibitem{advmix}
Sven Gowal, Chongli Qin, Po-Sen Huang, Taylan Cemgil, Krishnamurthy Dvijotham,
  Timothy Mann, and Pushmeet Kohli.
\newblock Achieving robustness in the wild via adversarial mixing with
  disentangled representations.
\newblock In {\em Proceedings of the IEEE/CVF Conference on Computer Vision and
  Pattern Recognition}, pages 1211--1220, 2020.

\bibitem{gu2020image}
Jinjin Gu, Yujun Shen, and Bolei Zhou.
\newblock Image processing using multi-code gan prior.
\newblock In {\em Proceedings of the IEEE/CVF Conference on Computer Vision and
  Pattern Recognition}, pages 3012--3021, 2020.

\bibitem{PCA}
Erik H{\"a}rk{\"o}nen, Aaron Hertzmann, Jaakko Lehtinen, and Sylvain Paris.
\newblock Ganspace: Discovering interpretable gan controls.
\newblock {\em arXiv preprint arXiv:2004.02546}, 2020.

\bibitem{huang2017arbitrary}
Xun Huang and Serge Belongie.
\newblock Arbitrary style transfer in real-time with adaptive instance
  normalization.
\newblock In {\em Proceedings of the IEEE International Conference on Computer
  Vision}, pages 1501--1510, 2017.

\bibitem{jalal2017robust}
Ajil Jalal, Andrew Ilyas, Constantinos Daskalakis, and Alexandros~G Dimakis.
\newblock The robust manifold defense: Adversarial training using generative
  models.
\newblock {\em arXiv preprint arXiv:1712.09196}, 2017.

\bibitem{kanbak2018geometric}
Can Kanbak, Seyed-Mohsen Moosavi-Dezfooli, and Pascal Frossard.
\newblock Geometric robustness of deep networks: analysis and improvement.
\newblock In {\em Proceedings of the IEEE Conference on Computer Vision and
  Pattern Recognition}, pages 4441--4449, 2018.

\bibitem{kannan2018adversarial}
Harini Kannan, Alexey Kurakin, and Ian Goodfellow.
\newblock Adversarial logit pairing.
\newblock {\em arXiv preprint arXiv:1803.06373}, 2018.

\bibitem{karras2019style}
Tero Karras, Samuli Laine, and Timo Aila.
\newblock A style-based generator architecture for generative adversarial
  networks.
\newblock In {\em Proceedings of the IEEE conference on computer vision and
  pattern recognition}, pages 4401--4410, 2019.

\bibitem{karras2020analyzing}
Tero Karras, Samuli Laine, Miika Aittala, Janne Hellsten, Jaakko Lehtinen, and
  Timo Aila.
\newblock Analyzing and improving the image quality of stylegan.
\newblock In {\em Proceedings of the IEEE/CVF Conference on Computer Vision and
  Pattern Recognition}, pages 8110--8119, 2020.

\bibitem{kurakin2016adversarial}
Alexey Kurakin, Ian Goodfellow, and Samy Bengio.
\newblock Adversarial machine learning at scale.
\newblock {\em arXiv preprint arXiv:1611.01236}, 2016.

\bibitem{lipton2017precise}
Zachary~C Lipton and Subarna Tripathi.
\newblock Precise recovery of latent vectors from generative adversarial
  networks.
\newblock {\em arXiv preprint arXiv:1702.04782}, 2017.

\bibitem{luo2017learning}
Junyu Luo, Yong Xu, Chenwei Tang, and Jiancheng Lv.
\newblock Learning inverse mapping by autoencoder based generative adversarial
  nets.
\newblock In {\em International Conference on Neural Information Processing},
  pages 207--216. Springer, 2017.

\bibitem{madry2017towards}
Aleksander Madry, Aleksandar Makelov, Ludwig Schmidt, Dimitris Tsipras, and
  Adrian Vladu.
\newblock Towards deep learning models resistant to adversarial attacks.
\newblock {\em arXiv preprint arXiv:1706.06083}, 2017.

\bibitem{moosavi2019robustness}
Seyed-Mohsen Moosavi-Dezfooli, Alhussein Fawzi, Jonathan Uesato, and Pascal
  Frossard.
\newblock Robustness via curvature regularization, and vice versa.
\newblock In {\em Proceedings of the IEEE Conference on Computer Vision and
  Pattern Recognition}, pages 9078--9086, 2019.

\bibitem{odena2017conditional}
Augustus Odena, Christopher Olah, and Jonathon Shlens.
\newblock Conditional image synthesis with auxiliary classifier gans.
\newblock In {\em International conference on machine learning}, pages
  2642--2651, 2017.

\bibitem{papernot2016distillation}
Nicolas Papernot, Patrick McDaniel, Xi Wu, Somesh Jha, and Ananthram Swami.
\newblock Distillation as a defense to adversarial perturbations against deep
  neural networks.
\newblock In {\em 2016 IEEE Symposium on Security and Privacy (SP)}, pages
  582--597. IEEE, 2016.

\bibitem{perarnau2016invertible}
Guim Perarnau, Joost Van De~Weijer, Bogdan Raducanu, and Jose~M {\'A}lvarez.
\newblock Invertible conditional gans for image editing.
\newblock {\em arXiv preprint arXiv:1611.06355}, 2016.

\bibitem{qiu2019semanticadv}
Haonan Qiu, Chaowei Xiao, Lei Yang, Xinchen Yan, Honglak Lee, and Bo Li.
\newblock Semanticadv: Generating adversarial examples via
  attribute-conditional image editing.
\newblock {\em arXiv preprint arXiv:1906.07927}, 2019.

\bibitem{simonyan2014very}
Karen Simonyan and Andrew Zisserman.
\newblock Very deep convolutional networks for large-scale image recognition.
\newblock {\em arXiv preprint arXiv:1409.1556}, 2014.

\bibitem{song2018constructing}
Yang Song, Rui Shu, Nate Kushman, and Stefano Ermon.
\newblock Constructing unrestricted adversarial examples with generative
  models.
\newblock In {\em Advances in Neural Information Processing Systems}, pages
  8312--8323, 2018.

\bibitem{szegedy2016rethinking}
Christian Szegedy, Vincent Vanhoucke, Sergey Ioffe, Jon Shlens, and Zbigniew
  Wojna.
\newblock Rethinking the inception architecture for computer vision.
\newblock In {\em Proceedings of the IEEE conference on computer vision and
  pattern recognition}, pages 2818--2826, 2016.

\bibitem{szegedy2013intriguing}
Christian Szegedy, Wojciech Zaremba, Ilya Sutskever, Joan Bruna, Dumitru Erhan,
  Ian Goodfellow, and Rob Fergus.
\newblock Intriguing properties of neural networks.
\newblock {\em arXiv preprint arXiv:1312.6199}, 2013.

\bibitem{vasiljevic2016examining}
Igor Vasiljevic, Ayan Chakrabarti, and Gregory Shakhnarovich.
\newblock Examining the impact of blur on recognition by convolutional
  networks.
\newblock {\em arXiv preprint arXiv:1611.05760}, 2016.

\bibitem{wang2018transferring}
Yaxing Wang, Chenshen Wu, Luis Herranz, Joost van~de Weijer, Abel
  Gonzalez-Garcia, and Bogdan Raducanu.
\newblock Transferring gans: generating images from limited data.
\newblock In {\em Proceedings of the European Conference on Computer Vision
  (ECCV)}, pages 218--234, 2018.

\bibitem{xiao2018generating}
Chaowei Xiao, Bo Li, Jun-Yan Zhu, Warren He, Mingyan Liu, and Dawn Song.
\newblock Generating adversarial examples with adversarial networks.
\newblock {\em arXiv preprint arXiv:1801.02610}, 2018.

\bibitem{xie2019feature}
Cihang Xie, Yuxin Wu, Laurens van~der Maaten, Alan~L Yuille, and Kaiming He.
\newblock Feature denoising for improving adversarial robustness.
\newblock In {\em Proceedings of the IEEE Conference on Computer Vision and
  Pattern Recognition}, pages 501--509, 2019.

\bibitem{yu2015lsun}
Fisher Yu, Ari Seff, Yinda Zhang, Shuran Song, Thomas Funkhouser, and Jianxiong
  Xiao.
\newblock Lsun: Construction of a large-scale image dataset using deep learning
  with humans in the loop.
\newblock {\em arXiv preprint arXiv:1506.03365}, 2015.

\bibitem{zhang2017mixup}
Hongyi Zhang, Moustapha Cisse, Yann~N Dauphin, and David Lopez-Paz.
\newblock mixup: Beyond empirical risk minimization.
\newblock {\em arXiv preprint arXiv:1710.09412}, 2017.

\bibitem{zhao2020differentiable}
Shengyu Zhao, Zhijian Liu, Ji Lin, Jun-Yan Zhu, and Song Han.
\newblock Differentiable augmentation for data-efficient gan training.
\newblock {\em arXiv preprint arXiv:2006.10738}, 2020.

\bibitem{zhu2016generative}
Jun-Yan Zhu, Philipp Kr{\"a}henb{\"u}hl, Eli Shechtman, and Alexei~A Efros.
\newblock Generative visual manipulation on the natural image manifold.
\newblock In {\em European conference on computer vision}, pages 597--613.
  Springer, 2016.

\end{thebibliography}
}

\end{document}